\documentclass[10pt,journal,compsoc]{IEEEtran}
\usepackage{times}
\usepackage{epsfig}
\usepackage{graphicx}
\usepackage{amsmath}
\usepackage{amssymb}
\usepackage{caption}
\usepackage{bbding}
\usepackage{color}

\ifCLASSOPTIONcompsoc
\usepackage[nocompress]{cite}
\else
\usepackage{cite}
\fi

\ifCLASSINFOpdf
\else
\fi

\begin{document}
	%
	\title{LSC-GAN: Latent Style Code Modeling for Continuous Image-to-image Translation}
	%
	%
	%
	%
	
	\author{Qiusheng~Huang, Xueqi~Hu,~Li~Sun,~\IEEEmembership{Member,~IEEE,}
		and~Qingli~Li,~\IEEEmembership{Senior Member,~IEEE}
		\IEEEcompsocitemizethanks{\IEEEcompsocthanksitem Q. Huang, L. Sun and Q. Li are with the Shanghai Key Laboratory of Multidimensional Information Processing, East China Normal University, Shanghai 200241, China. L. Sun is the corresponding author. \protect\\
			E-mail: sunli@ee.ecnu.edu.cn 
	} }
	
	\markboth{Journal of \LaTeX\ Class Files,~Vol.~14, No.~8, August~2015}%
	{Shell \MakeLowercase{\textit{et al.}}: Bare Demo of IEEEtran.cls for Computer Society Journals}
	
	\IEEEtitleabstractindextext{%
		\begin{abstract}
			Image-to-image (I2I) translation is usually carried out among discrete domains. However, image domains, often corresponding to a physical value, are usually continuous. In other words, images gradually change with the value, and there exists no obvious gap between different domains. This paper intends to build the model for I2I translation among continuous varying domains. We first divide the whole domain coverage into discrete intervals, and explicitly model the latent style code for the center of each interval. To deal with continuous translation, we design the editing modules, changing the latent style code along two directions. These editing modules help to constrain the codes for domain centers during training, so that the model can better understand the relation among them. To have diverse results, the latent style code is further diversified with either the random noise or features from the reference image, giving the individual style code to the decoder for label-based or reference-based synthesis. Extensive experiments on age and viewing angle translation show that the proposed method can achieve high-quality results, and it is also flexible for users. \footnote{The code will be released in https://github.com/huangqiusheng/LSC-GAN-Latent-Style-Code-Modeling-for-Continuous-Image-to-image-Translation.}
		\end{abstract}
		
		}
	
	\twocolumn[{
		\renewcommand\twocolumn[1][]{#1}
		\maketitle
		\begin{center}
			\centering 
			\vspace{-1.0cm}
			\setlength{\belowcaptionskip}{1cm}
			\includegraphics[width=1.00\textwidth]{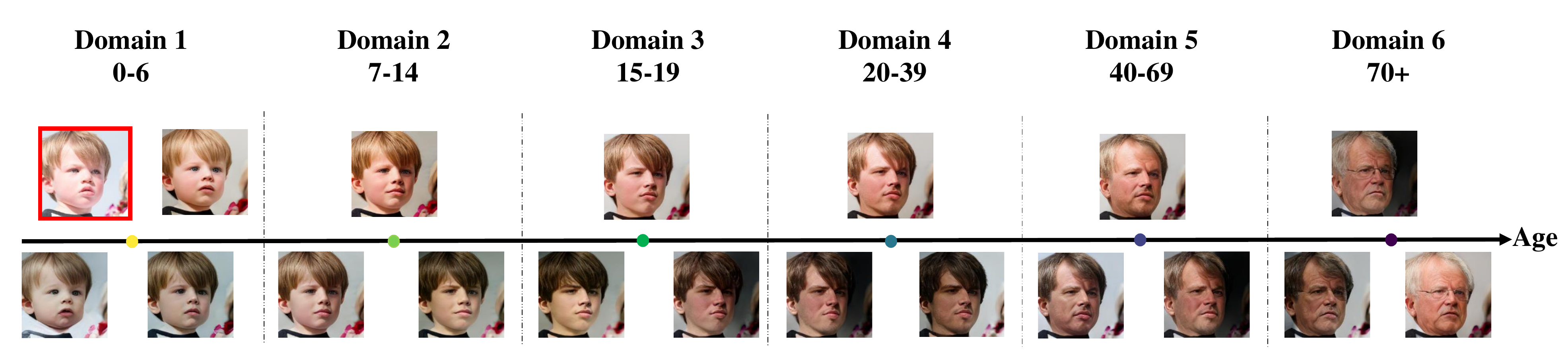}
			\captionof{figure}{\textbf{
					Diverse face aging 
					results in each domain.
				}The image in the red box is the source, and the others are the label-based synthesis 
				from our model, 
				which are obtained by 
				translating the source to that domain. Note that images in the same domain show obviously different ages, demonstrating that our model can give attribute-relevant diversity within the same domain. } 
			\label{fig:fig0}
		\end{center}
	}]
	
	\IEEEdisplaynontitleabstractindextext
	
	\IEEEpeerreviewmaketitle

	\IEEEraisesectionheading{\section{Introduction}\label{sec:introduction}}
	\IEEEPARstart{G}{enerative} model, particularly the technique of Generative Adversarial Network (GAN) \cite{brock2018large,goodfellow2014generative,odena2017conditional}, is well-developed in recent year. It can synthesize high-resolution realistic images by mimicking the data distribution, through the adversarial training of two deep Convolutional Neural Networks (CNN), namely the generator G and discriminator D. The G maps the randomly sampled noise vector into the color image, which is inspected by D so that it has the aligned output distribution with real data. Conditional GAN \cite{2014Conditional, pmlr-v70-odena17a} is able to synthesize the image which fulfills the requirement of the given label. Usually, the label indicates the discrete type of the image, which is encoded by a one-hot vector. It is adopted by both G and D, so that the conditional data distribution for a specific class can be learned by G. 
	
	Due to the success of GAN and cGAN, direct image-to-image (I2I) translation across different domains can be accomplished by a generator G in cGAN. It aims to learn the mapping function, which changes the source domain input image into the target domain. The result should take the content of input. At the same time, it also needs to satisfy the target domain requirement. In practice, labels reflecting a particular attribute are often not discrete, but they are more appropriately described by a continuous value, which varies within a certain range, \emph{e.g.}, the age or the viewing angle of a face. Such labels define the continuous image domains, which require G being able to be controlled so that its output naturally changes among the domains. In image generation, GANs \cite{karras2019style, karras2020analyzing} usually achieve this through "morphing". By interpolating the input noise (or label vector) from $z_1$ to $z_2$, the intermediate results from $\text{G}(z_1)$ to $\text{G}(z_2)$ change in the smooth way, showing that $\text{G}$ has the strong ability to map a noise vector into an image. However, the quality and the domain conformity of the intermediate results can not be ensured. 
	
	In I2I translation, existing techniques lack the special designs to deal with the continuous domains. CycleGAN \cite{zhu2017unpaired}, StarGAN \cite{choi2018stargan} and StarGAN-v2 \cite{choi2020stargan} all focus on the translations among discrete domains, which are either caused by a single or multiple attributes. Although, by simply dividing the whole domain range into several discrete intervals, these models can still perform I2I translation, the relation among different domains are ignored during training, and the domain conformity of the result is not accurate enough. Moreover, CycleGAN and StarGAN are label-based, which can only translate the source image by providing the target label. They are impossible to output the diverse synthesis. StarGAN-v2 can support both the label- and reference-based translation, and it synthesizes different results for the same source in the target domain. But these results only show the attribute-irrelevant diversity, such as the different lightness and backgrounds. It can not slightly modify the desired attribute within the same domain, \emph{i.e.}, changing the age from 5 to less than 3. In summary, these models lack the fine control ability for the continuous domain. 
	
	This paper aims for the continuous I2I translation, \emph{i.e.}, editing the facial age or viewing angle. We intentionally consider the relation among the predefined discrete domains. Similar to StarGAN-v2, we achieve both the label- and the reference-based synthesis, and the model is able to output the diverse images, all fulfilling the target domain requirement. To accurately capture the characteristic of each domain interval, we first explicitly build the representative style codes of the domain centers. Then they are encoded together with the random noise or reference image to give diverse and specific styles. To make the model understand the domain relations, we design the two types of latent code editing modules, which are capable of changing the code along two opposite directions (\emph{e.g.} young and old directions), so that the style code from one domain can be translated into its two neighbouring domains. These modules help to constrain the distance among domain centers during training. Moreover, they are utilized to edit the style code, making the consistent change in a specific direction in the inference stage. Note that with these modules, the latent style code can even go beyond the initial domain coverage. So the model becomes flexible for the user during the inference. 
	
	The contribution of this paper lies in following aspects. 
	\begin{itemize}
		\item
		We propose a generic I2I translation model which is particularly suitable for continuous domains. By modeling the latent style code for each domain center, our method preserves the common domain styles well. They are further diversified by either the noises or reference images, which give the corresponding label- or reference-based translations, respectively.
		\item 
		We design the latent style editing module to convert the code for domain center into other domains, and set up the relation constraint among them. The module can be further adopted to edit the image consistently in one direction during the inference stage.
		\item 
		A plenty of experiments are performed on two different datasets with continuous domain labels, FFHQ-Aging and Multi-PIE. In the former, our model translates the age of the face into an arbitrary age group. While in the latter, it is able to edit the viewing angle of the face. Both of them show the superiority of our proposed model.
	\end{itemize}
	
	\begin{figure*}[h]
		\centering
		\includegraphics[width=0.8\textwidth]{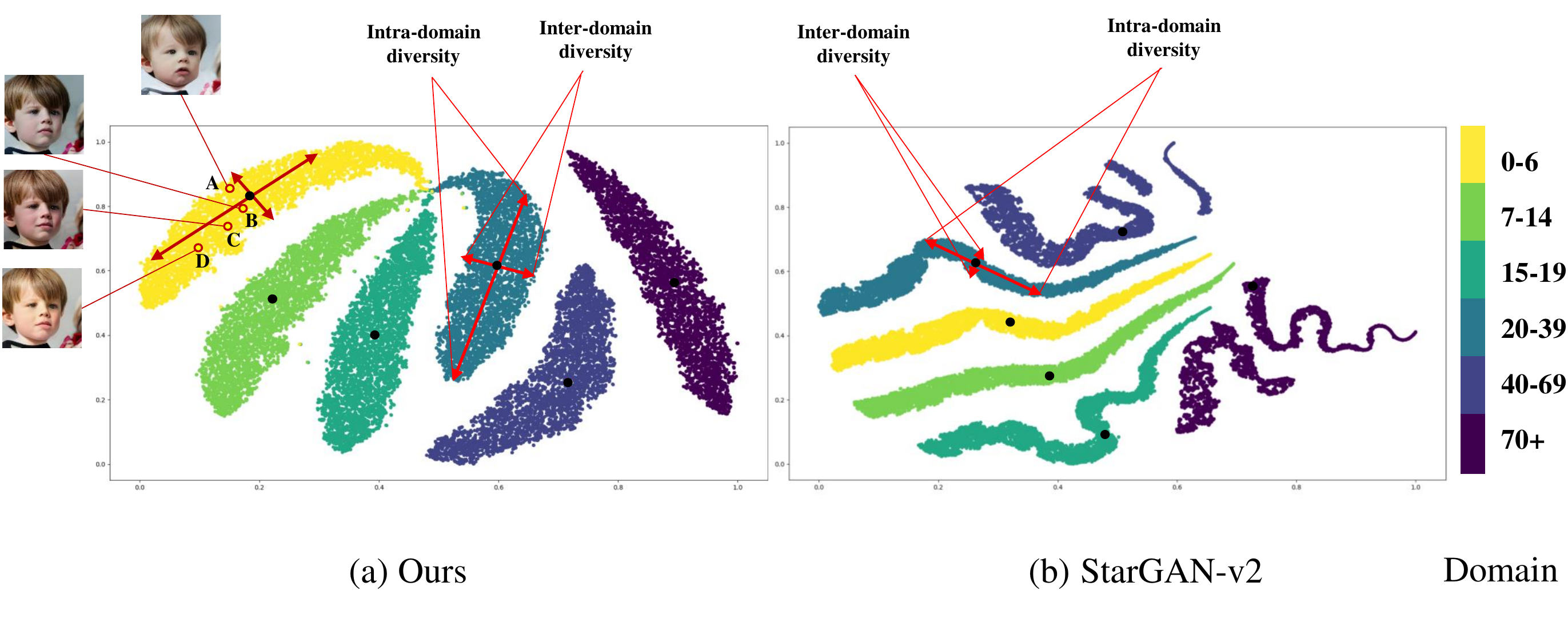} 
		\caption{\textbf{TSNE visualizations on the latent style codes} $s_l$. After training, we have several domain center $c$. Together with different sampling noises, they are encoded into $s_l$. Here we sample $5,000$ noise vectors, leading to the same amount of $s_l$ for each domain. Within a domain, we use red arrows to indicate different directions of variation, representing the intra- and intra-domain diversity, respectively. We calculate the average of all the code in each domain and display it as a black dot. }
		\label{fig:tsne}
	\end{figure*}
	\newcommand{\tabincell}[2]{\begin{tabular}{@{}#1@{}}#2\end{tabular}}  
	\begin{table*}[h]
		\begin{center}
			\scalebox{1.0}{
				\setlength{\tabcolsep}{1.5mm}{
					\begin{tabular}{ l l l c c c c c c}
						\hline
						&Method && \tabincell{c}{Latent-guided \\ synthesis} &\tabincell{c}{Reference-guided \\ synthesis}
						&\tabincell{c}{Intra-domain \\ diversity}
						&\tabincell{c}{Inter-domain \\ diversity}
						&\tabincell{c}{Continuous mapping \\ from domain to domain}
						&\tabincell{c}{Beyond domain \\ limit}\\
						\hline
						\hline
						&StarGAN-v2 \cite{choi2020stargan}&& \Checkmark&\Checkmark&\Checkmark&\XSolidBrush&\XSolidBrush&\XSolidBrush\\
						&DivCo \cite{liu2021divco}&& \Checkmark&\Checkmark&\Checkmark&\XSolidBrush&\XSolidBrush&\XSolidBrush\\
						&DRIT++ \cite{lee2020drit++}&& \Checkmark&\Checkmark&\Checkmark&\XSolidBrush&\XSolidBrush&\XSolidBrush\\
						&CVAE-GAN \cite{bao2017cvae}&& \Checkmark&\XSolidBrush&\XSolidBrush&\XSolidBrush&\XSolidBrush&\XSolidBrush\\
						&DLOW \cite{gong2019dlow}&& \Checkmark&\XSolidBrush&\XSolidBrush&\XSolidBrush&\Checkmark&\XSolidBrush\\
						&LIFE \cite{or2020lifespan}&& \Checkmark&\XSolidBrush&\XSolidBrush&\XSolidBrush&\XSolidBrush&\XSolidBrush\\
						&StarGAN \cite{choi2018stargan}&& \Checkmark&\XSolidBrush&\XSolidBrush&\XSolidBrush&\XSolidBrush&\XSolidBrush\\
						&STGAN \cite{liu2019stgan}&& \Checkmark&\XSolidBrush&\XSolidBrush&\XSolidBrush&\XSolidBrush&\XSolidBrush\\
						&SAM \cite{alaluf2021only}&& \Checkmark&\XSolidBrush&\XSolidBrush&\XSolidBrush&\XSolidBrush&\XSolidBrush\\
						&IPCGAN \cite{wang2018face}&& \Checkmark&\XSolidBrush&\XSolidBrush&\XSolidBrush&\XSolidBrush&\XSolidBrush\\
						&VI-GAN \cite{xu2019view}&& \Checkmark&\XSolidBrush&\XSolidBrush&\XSolidBrush&\Checkmark&\XSolidBrush\\
						&CD-VAE \cite{yin2020novel}&& \Checkmark&\XSolidBrush&\XSolidBrush&\XSolidBrush&\Checkmark&\XSolidBrush\\
						
						\hline
						\hline
						&Ours&&\Checkmark&\Checkmark&\Checkmark&\Checkmark&\Checkmark&\Checkmark\\
						\hline
				\end{tabular}}
			}
		\end{center}
		\caption{\textbf{Comparisons with approaches in I2I translation.} }
		\label{tab:1}
	\end{table*}
	\section{Related Work}
	\subsection{I2I translations between two domains}
	The topic of I2I translation has drawn researchers' great attention in recent years. It is first proposed in Pix2Pix \cite{isola2017image}, and extended to Pix2PixHD \cite{wang2018high} for translating the high-resolution image. The generator G of both models are built by the auto-encoder, with the first half reducing the spatial resolution and encoding the content of a source domain image $x$, while the second half enlarging the size and translating into the target image. These models need the paired data during training, which means the output from G are directly constrained by the $L_1$ or $L_2$ distance with the target image. CycleGAN \cite{zhu2017unpaired}, DualGAN \cite{yi2017dualgan} and DiscoGAN \cite{kim2017learning} relax the assumption to the unpaired data. They build $\text{G}_{X\rightarrow Y}$ and $\text{G}_{Y\rightarrow X}$ for two translating directions, and set up the consistency between $x$ and $\text{G}_{Y\rightarrow X}(\text{G}_{X\rightarrow Y}(x))$. Note that these models are only able to give single-modal results, and they implicitly edit the 
	image by the target domain label.
	
	To obtain diverse results, one way is to change the backbone 
	$\text{G}$ to 
	VAE \cite{kingma2013auto,bao2017cvae,zheng2019disentangling,yin2020novel}, in which the latent code from the encoder is 
	sampled from the posterior distribution determined by the input $x$. Another solution for diversifying the results is to incorporate a reference image in G, therefore, the same content can be combined with different references, leading to various results \cite{liu2017unsupervised,huang2018multimodal,lee2018diverse,lee2020drit++}. Note that the synthesis needs to show a similar style with the reference. UNIT \cite{liu2017unsupervised} uses two VAEs, mapping images from different domains to a shared probabilistic space. Then the decoders directly translate the sampled latent code into either the original or the other domain. MUNIT \cite{huang2018multimodal} and DRIT \cite{lee2020drit++} explicitly disentangle between the content and the style code to 
	promote the diverse styles. They add an extra encoder, specifying a style code,  
	and inject it into G by AdaIN \cite{huang2017arbitrary}. 
	
	In addition, 
	training loss is also important for diverse results. MS-GAN \cite{Mao_2019_CVPR} and DivCo \cite{liu2021divco} design the loss to prevent the mode collapse in image generation. The former directly maximizes ratio of the pixel domain distance, with respect to the distance in the latent space. The latter adopts the similar idea through the contrastive learning, constraining the two close latent codes producing the similar visual results.
	
	\subsection{I2I translations among multiple domains}
	Previous models mainly deals with two domains. In practice, it is sometimes necessary to convert images among multiple domains. 
	StarGAN \cite{choi2018stargan} is the first work which achieves translation among multiple domains 
	through a single $G$. It uses the target label and the original image as two inputs, and outputs the label based synthesis, which is inspected by a discriminator. Similar to \cite{zhu2017unpaired}, it also adopts the cycle consistency to ensure $G$ to take the content from the input. Instead of the target label, AttGAN \cite{he2019attgan} employs the label difference as the input, and changes the image towards to the target domain. 
	ST-GAN \cite{liu2019stgan} designs a module to iteratively process the encoder features from different layers, and inject them into the multiple layers of the decoder. All these models realize the translation according to the specified label. They are deterministic, so can not give diverse results. 
	
	A common solution for diversity is to bring in noises in $G$. SMIT \cite{romero2019smit} extends StarGAN \cite{choi2018stargan} by concatenating a random noise vector with the target label, and then mapping them into a style vector, which influences the statistics of the feature in $G$. To prevent the noise being ignored, model parameters for processing the noise are not trained in SMIT. Similar with two domains task, adding an extra image as a reference is another option to diversify the synthesis.
	ELEGANT \cite{xiao2018elegant} exchanges and mixes the latent codes from two images. It consists of only a pair of connected encoder-decoder, which is able to either reconstruct the input image or edit the specified attribute to the target domain. HomoGAN \cite{chen2019homomorphic} learns interpolators between two latent codes from source and target images. Then it maps the interpolated codes into images, so that they exhibit the gradual change from the source to the target. Note that although the reference based model can give diverse result, it usually has the lower quality than label based model.
	
	To achieve high quality and diverse translations, StarGAN-v2 combines the label and reference based synthesis in a single model. It also has careful design on the training loss. Since StarGAN-v2 becomes the current SOTA model for I2I translation, we build our proposed model based on it, and give a brief introduction about it in the next section. 
	
	\subsection{Continuous domain I2I translation} The above works deal with discrete domains. However, the I2I translation for continuous domains has more applications. Here we focus on the facial age and viewing angle translation. Most works \cite{zhang2017age,wang2018face,he2019s2gan} for age edition divide the whole range into domain intervals, however, they lack special design for modeling the relation among them. 
	Moreover, they can only translate according to the target label, so the results lacks the diversity. Another problem is that these models are trained on the constrained data (e.g.frontal view and similar background), they have poor results on wild images. LIFE \cite{or2020lifespan} is the first model training on high-quality wild images. SAM \cite{alaluf2021only} uses a well-trained StyleGAN to produce high-quality age-edited images but still cannot produce diversity.
	
	Different from age edition, many works \cite{xu2019view,yin2020novel} for view angle edition considers the continuous mapping through the geometry relation between two domains. However, they only deal with the geometry related translation and lack the diverse results. 
	DLOW \cite{gong2019dlow} synthesizes transitional images between the two domains, in which a random scalar, ranging in $[0,1]$ and reflecting the domainness, is used as a pseudo label to control the distance between the intermediate and the source/target domains. Although DLOW achieves the continuous I2I translation, the intermediate domains are not determined by a physical value and do not have true data in their assumption. Therefore, the domain conformity is not guaranteed by real data. Moreover, it makes translation by the domainness, and lacks the ability for reference-based synthesis. 
	Tab.\ref{tab:1} lists the visual comparison between the above works and our proposed approach. 
	
	\begin{figure*}[h]
		\centering
		\includegraphics[width=0.8\textwidth]{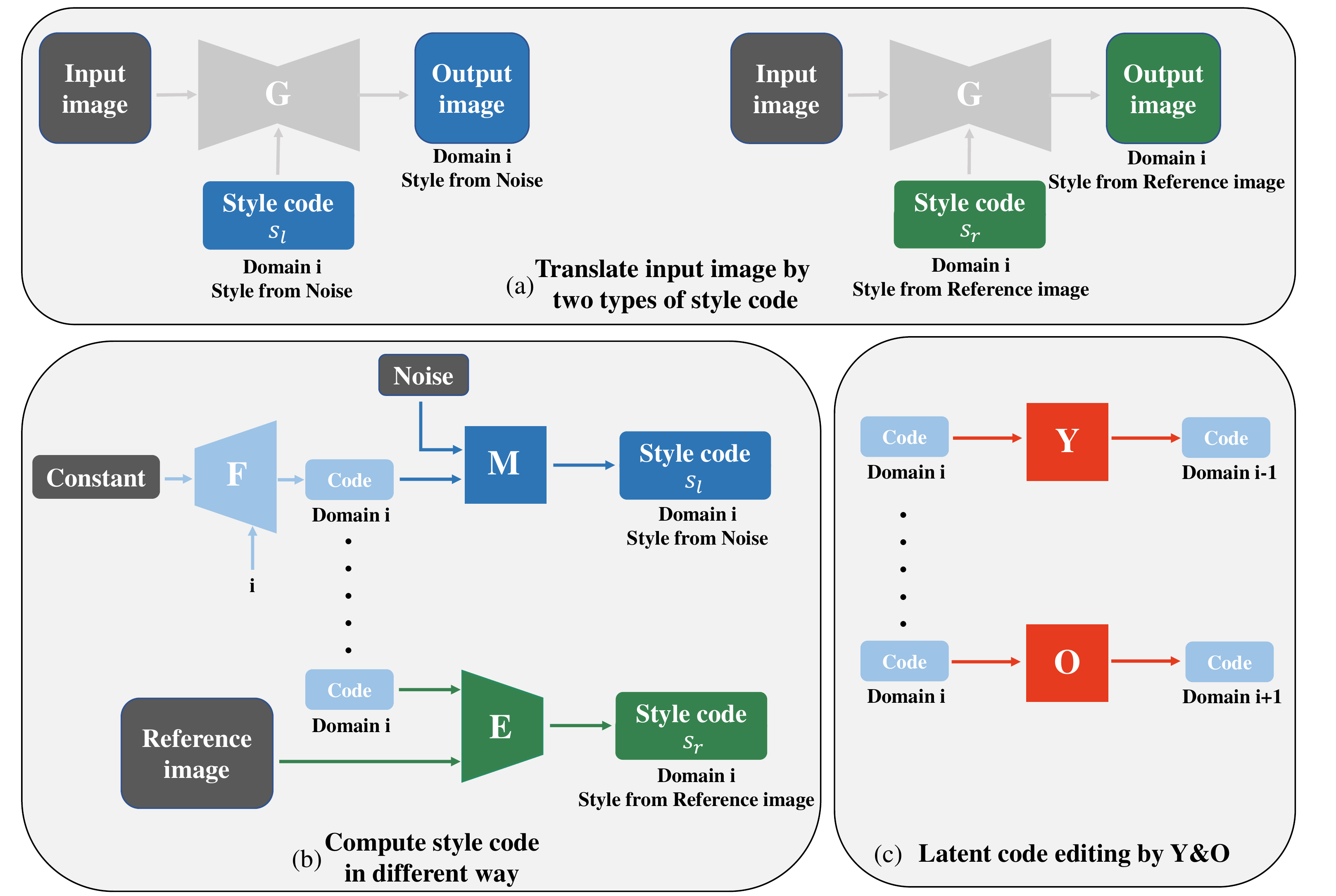}
		\caption{\textbf{Overview of our method.} (a) By feeding the auto-encoder G with two different types of style code ($s_l$ and $s_r$), translated image in $i$th domain with diverse styles can be obtained. The style codes are given to the same G with shared parameters. (b) Details of the computation of $s_l$ and $s_r$ for the $i$th domain. The latent codes for each domain are first output by the multi-branch F. Then they are further diversified to form $s_l$ or $s_r$ by the module M or E, with either the sampling noise vector or the reference image as its input. (c) The code from F can be edited by the module Y\&O along two opposite directions, so the $i$th domain code changes into its neighbouring domains, the $(i-1)$th or $(i+1)$th domain. 
		}
		\label{fig:fig1}\vspace{-0.5cm}
	\end{figure*}
	\section{Proposed Method}
	\subsection{Preliminaries on StarGAN-v2}
	\label{sec:s}
	Our work is based on StarGAN-v2 \cite{choi2020stargan}, which can realize both the label- and reference-based translation. It has an auto-encoder G as the backbone of the generator, and its feature in each layer is affected by the style vector injected from the side branch. 
	For the label-based synthesis, StarGAN-v2 uses a mapping network F to directly project a noise into a domain style code. F adopts a multi-branch structure, and the output 
	from the branch indicated by the label is further used. For the reference-based editing, it directly encodes a reference image into the style vector through an encoder E, 
	which is a CNN with 
	a multi-branch structure. When editing an image, the source is input into G, and the style code from F or E is injected into G 
	by AdaIN to get the 
	translated image. 
	
	In addition, StarGAN-v2 adopts 4 loss functions. $L_{adv}$ is an adversarial loss with gradient penalty \cite{NIPS2017_892c3b1c,2017Wasserstein} 
	to make the translated image real. 
	In order to make the style of the edited image close enough to that of the reference, and to retain the domain-invariant characteristics in the source image, 
	$L_{sty}$ and $L_{cyc}$ are employed. $L_{sty}$ reconstructs the style code of label-based synthesis by E, and $L_{cyc}$ is the cycle consistency computed on pixels. At the same time, $L_{ds}$ is introduced to encourage the 
	stronger diversity. For shorthand, $L_{Star}=L_{adv}+L_{sty}+L_{ds}+L_{cyc}$.
	
	\subsection{Proposed framework}
	Given a real image $x\in X$, and an arbitrary domain $y_t \in Y$, where $X$ 
	and $Y$ represent the sets of all images and 
	domains, respectively, we want to 
	edit $x$, making it have the style of the target domain $y_t$, 
	which can also be said that 
	$x$ is translated 
	to domain $y_t$. Traditional I2I translation usually assumes discrete domains, where $y_t$ with $t=1,2,\cdots,N$ represents $N$ discrete domains. Here, we mainly deal with the continuous domain. Particularly, the whole domain $Y$ is divided into $N$ discrete intervals, also indicated by $y_t$. The interval does not overlap with each other. With the subscript $t$ increases or decreases, the domain changes consistently in one direction. 
	
	For the continuous I2I translation, the 
	result is required to change according to $y_t$ in the consistent way, and it also exhibits diverse intra-domain and inter-domain features. 
	To achieve this, 
	we need first build 
	different 
	style vectors, and then train the generator 
	to use these style vectors for translation. 
	Fig. \ref{fig:tsne} compares the latent style code distribution between StarGAN-v2 and the proposed LSC-GAN.
	Fig. \ref{fig:fig1} describes the overview of the major blocks in the generator, and the two different pipelines for 
	synthesis. 
	
	\subsubsection{Auto-encoder G}Given the style 
	$s\in\{s_l, s_r\}$ belonging to the target domain $y_t 
	$, the role of G is to translate the source image $x$ to the domain $y_t$, 
	so that $\text{G}(x, s)$ gives diverse and appropriate results, conforming to $y_t$. 
	In our design, there are two types of style vectors, $s_l$ and $s_r$ for label- and reference-based synthesis, shown in Fig. \textcolor[rgb]{1,0,0}{\ref{fig:fig1}}a. Both of them 
	can express the desired style 
	of the target domain, 
	and they are the keys to enabling G to make the change on $x$ accordingly. 
	Note that we embed the style vector $s$ into G 
	through AdaIN \cite{huang2017arbitrary}, which affects the feature statistics in each layer of $\text{G}$.
	
	\subsubsection{Mapping network F for the center of domain interval} 
	To represent the domain interval center $c$, and set up the continuous relation among them, 
	we use the network F to learn the mapping $c_t=\text{F}(y_t)$, as is shown in Fig. \ref{fig:fig1}b. Similar to StarGAN-v2, a multi-branch network is adopted in 
	F. 
	Specifically, we utilize a total of $N$ branches at the output of F, 
	which form the 
	representative code $c_t$ 
	for each domain interval, and the parameters in the branch are not shared. 
	A constant vector is given as the input of F, 
	and 
	$N$ different $c$ can be obtained simultaneously 
	from the output of $F$. 
	Then we can select the code $c_t$, 
	according to $y_t$, for the further process.
	
	\subsubsection{Mapping network M for noise vector} 
	The purpose of 
	M is to 
	diversify the style 
	on the center 
	$c$, so as to map 
	it to the style vector $s_l=\text{M}(c_t,R)$ 
	in Fig. \ref{fig:fig1}b. Here $R$ is a randomly sampled noise vector. In fact, the code $c_t$ from $F$ expresses the 
	common characteristics of the domain $y_t$. Then we try to add 
	variation to $c_t$ by 
	$R$, so that 
	$s_l$ will increase the diversity of the final result on the premise that it still 
	has the basic characteristics of the 
	domain $y_t$. Particularly, we concatenate the 
	$R$ and the 
	$c_t$ output from F in the channel dimension and feed them into 
	M to generate the style vector $s_l$.
	
	\subsubsection{Style encoder E for the reference image} 
	
	Similar with StarGAN-v2, our 
	method also accepts 
	the reference image for modeling the style, since the dual-drive greatly encourages the diverse editing results. 
	The module E in Fig. \ref{fig:fig1}b is used to encode the reference input $x_t$. 
	The intermediate 
	features of $x_t$ are 
	encoded with the domain 
	center $c_t$, and then they are together mapped into the reference style 
	$s_r=\text{E}(x_t, c_t)$, 
	further exploited by the auto-encoder G for the reference-based synthesis. $s_r$ not only ensures the translation to be in the target domain, 
	but also make it have the 
	similar style 
	with the input $x_t$. Functionally, E can be compared with M. 
	
	\subsubsection{
		Latent code editing module Y\&O} 
	To model the relation of the 
	center $c$ among different domain intervals, we build two editing modules Y and O to change $c$ along two opposite directions consistently, as is shown in Fig. \ref{fig:fig1}c. They have the same structure but do not share the model parameters. Given a domain code $c_t$ generated by the network F, where $t=1,2,\cdots N$, the module O is expected to translate $c_t$ into its neighbouring domain $c_{t+1}$. This means that O($c_t$) becomes close with $c_{t+1}$. Similarly, the module Y changes $c$ in the opposite direction, and it is expected that Y($c_t$) is close with the other neighbour $c_{t-1}$. Note that edited features by Y\&O do not participate the generation process during training. They are only employed to restrict $c_t$ of each domain. We will illustrate the loss designed for them in section \ref{sec:obj}. Moreover, Y\&O can be conveniently used to translate the source, even beyond the initial domain coverage during inference.
	
	\subsubsection{Discriminator D} 
	Like StarGAN-v2, we adopt a multi-task discriminator to conduct adversarial training on each domain separately.
	Given an image $x$ and the domain $y$ it belongs to, D learns to distinguish whether $x$ is a real or fake image 
	during training.
	
	\subsection{Training objectives}\label{sec:obj}
	Since our network is based on StarGAN-v2, we use the similar training loss $L_{Star}$.  
	Here we only illustrate the extra terms. 
	Let $\{x,y_i\}$ be the source input and its corresponding domain label, and $\{x_t,y_t\}$ be the reference image with the target domain label, where $i,t=1,2,\cdots,N$, and $y$ (including $y_i$ and $y_t$) is the integer ranging from 
	$[1,N]$, increasing or decreasing its value by 1 as the subscript changes to the right or left neighbour. Note that the division of domain coverage 
	is based on the actual physical value, and 
	domains are organized to keep the internal orders. 
	Following 3 loss terms are designed, namely $L_{cdc}$, $L_{tam}$ and $L_{ccc}$, to model the 
	center $c$ 
	for each domain interval. 
	$c_i=\mathrm{F}(y_i)$ and $c_t=\mathrm{F}(y_t)$ are centers for the source and target domain, respectively.
	\subsubsection{Continuous domain constraint} As is assumed above, there are two types of latent editing modules, named Y and O, to translate the code $c_i$ into different domains specified by the target label $y_t$.  
	Here Y\&O are supposed to perform successive editing on $c_i$, changing it in one direction by recursively modifying the previous output. As is shown in (\ref{2}), $\mathrm{Y}^k/\mathrm{O}^k$ means to call module Y/O $k$ times, repeatedly. $\stackrel{\rightarrow}{c}_t$ indicates the target domain $y_t$ has larger value than $y_i$, so the module O is applied on $c_i$ with $y_t-y_i$ times. On the other hand, $\stackrel{\leftarrow}{c}_t$ implies $y_i>y_o$, so the module Y is used to edit $c_i$ with several times. 
	
	\begin{equation}
		\label{2}
		\begin{aligned}
			&\stackrel{\rightarrow}{c}_t=\mathrm{O}^{y_t-y_i}(c_i)
			&\stackrel{\leftarrow}{c}_t=\mathrm{Y}^{y_i-y_t}(c_i)
		\end{aligned}
	\end{equation}
	Given a pair of the source and target label $y_i$ and $y_t$, we first choose to edit $c_i$ by the module O or Y, leading to $\stackrel{\rightarrow}{c}_t$ or $\stackrel{\leftarrow}{c}_t$. Then, the loss $L_{cdc}$ is computed according to (\ref{1}). 
	\begin{eqnarray}
		\label{1}
		L_{cdc} =
		\begin{cases}
			| c_t - \stackrel{\rightarrow}{c}_t |   & y_t-y_i>0 \\
			| c_t - \stackrel{\leftarrow}{c}_t |   & y_t-y_i<0 \\
			0  & y_t-y_i=0
		\end{cases}
	\end{eqnarray}
	Basically, there are 3 different cases. The first and second item utilize $\stackrel{\rightarrow}{c}_t$ and $\stackrel{\leftarrow}{c}_t$, which are further depended on O and Y module, respectively. $L_{cdc}$ is employed to train the parameters in O, Y, M and F, building the relation among $c_t$. 
	
	\subsubsection{Triplet loss with adaptive margin}
	We adopt a triplet loss to further ensure the edited latent code ($\stackrel{\rightarrow}{c}_t$ or $\stackrel{\leftarrow}{c}_t$) in the target domain, 
	as is shown in (\ref{3}). Here $\stackrel{\rightarrow}{c}_t$ or $\stackrel{\leftarrow}{c}_t$ is the anchor, $c_t$ is the positive which is directly computed from F, and $c_j$ is the negative. $L_{tam}$ computes the Euclidean distance $d(\stackrel{\rightarrow}{c}_t,c_t)$ or $d(\stackrel{\leftarrow}{c}_t,c_t)$ between the anchor and positive, and compares it with the $d(\stackrel{\rightarrow}{c}_t, c_j)$ or $d(\stackrel{\leftarrow}{c}_t, c_j)$ between the anchor and negative.
	\begin{small}
		\begin{eqnarray}
			\label{3}
			L_{tam} =
			\begin{cases}
				\max_j(d(\stackrel{\rightarrow}{c}_t,c_t) -d(\stackrel{\rightarrow}{c}_t, c_j) + m, 0)  & y_t-y_i>0 \\
				\max_j(d(\stackrel{\leftarrow}{c}_t,c_t) - d(\stackrel{\leftarrow}{c}_t, c_j) + m, 0)  & y_t-y_i<0 \\
				0  & y_t-y_i=0 
			\end{cases}
		\end{eqnarray}
	\end{small}
	$c_j$ is the latent code of an arbitrary domain $y_j$, computed by F. Hence, $L_{tam}$ is determined by the hardest negative through $\max(\cdot)$ on the index $j$. $m$ is the adaptive margin to adjust the distance between two domains, and we set $m=|t - j|$. In other words, the greater the distance between $y_t$ and $y_j$, the greater the value of margin. 
	
	\subsubsection{Cycle continuous consistency loss} 
	To further emphasize the role of the module Y\&O and stabilize the latent code $c$ for the center of each domain interval, we use the following loss function in (\ref{4}).
	\begin{eqnarray}
		\label{4}
		L_{ccc} =
		\begin{cases}
			| c_i - \mathrm{Y}^{y_t-y_i}(\stackrel{\rightarrow}{c}_t) |   & y_t-y_i>0 \\
			| c_i - \mathrm{O}^{y_i-y_t}(\stackrel{\leftarrow}{c}_t) |   & y_t-y_i<0 \\
			0  & y_t-y_i=0 \\
		\end{cases}
	\end{eqnarray}
	Here $L_{ccc}$ is the cycle continuous consistency loss obtained by editing $\stackrel{\rightarrow}{c}_t$ (or $\stackrel{\leftarrow}{c}_t$) in the opposite direction defined by $\mathrm{Y}^{y_t-y_i}$ (or $\mathrm{O}^{y_i-y_t}$).
	\subsubsection{Full objective} 
	Finally, we train our $\text{M}$, $\text{F}$, $\text{Y\&O}$, $\text{E}$, $\text{G}$ and $\text{D}$, to minimize following objectives.
	\begin{equation}
		\label{5}
		\begin{aligned}
			&L_{\mathrm{EDG}} = L_{Star}\\
			&L_{\mathrm{FMYO}} = L_{Star} + \lambda_{cdc}L_{cdc} + \lambda_{tam}L_{tam} + \lambda_{ccc}L_{ccc} 
		\end{aligned}
	\end{equation}
	where $\lambda_{cdc}$, $\lambda_{tam}$ and $\lambda_{ccc}$ are hyper-parameters for each term. $L_{Star}$ (\ref{sec:s}) is the loss functions used by StarGAN-v2.
	
	\begin{figure*}
		\centering
		\includegraphics[width=0.8\textwidth]{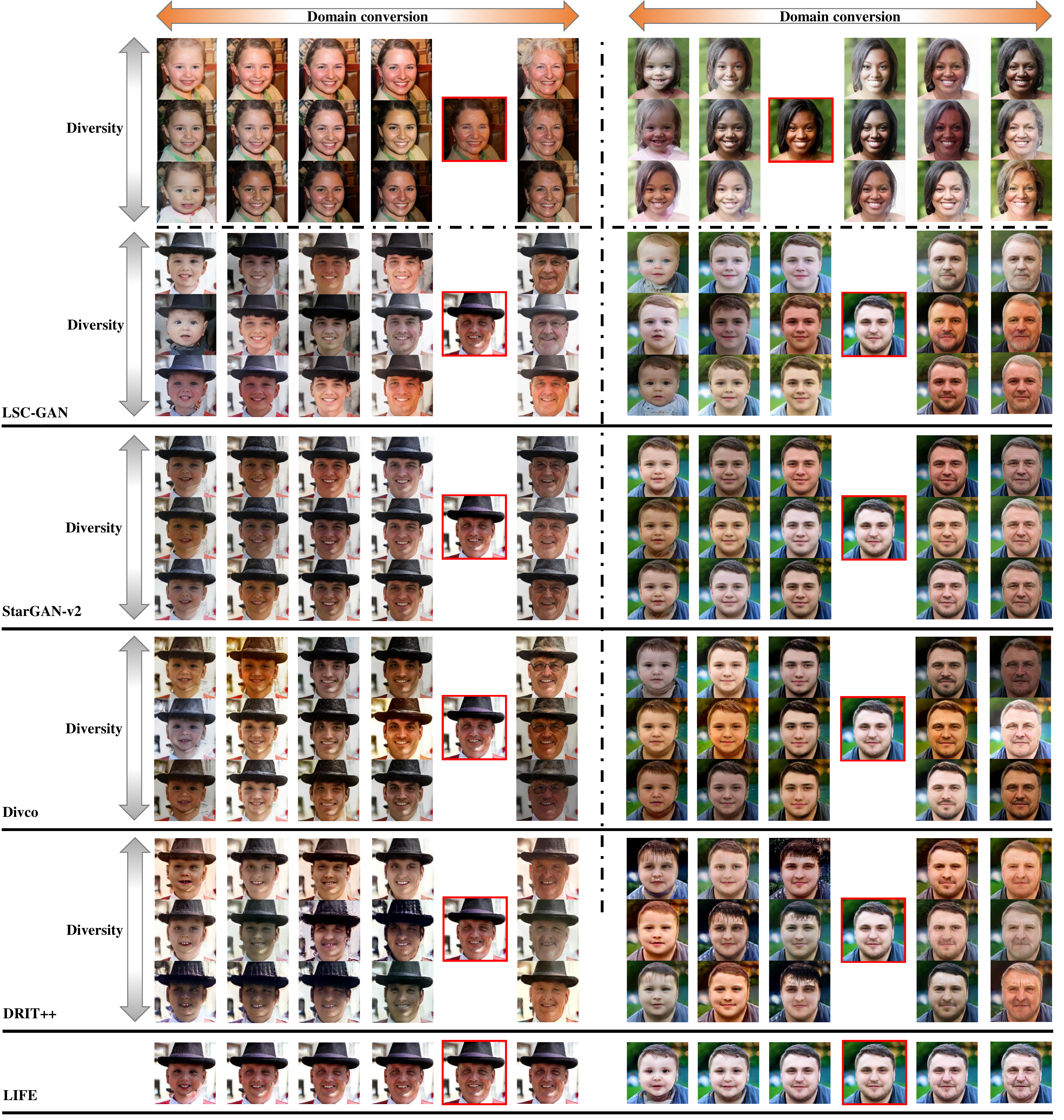}
		\caption{\textbf{
				Label-based comparison on FFHQ-Aging.} The real images in red box are input into the model for translation. To show the diversity, 3 different noises are sampled for the same domain. We also compare 2 sets of results with other works.}
		\label{fig:latent1}
	\end{figure*}
	
	\begin{figure*}
		\centering
		\includegraphics[width=0.8\textwidth]{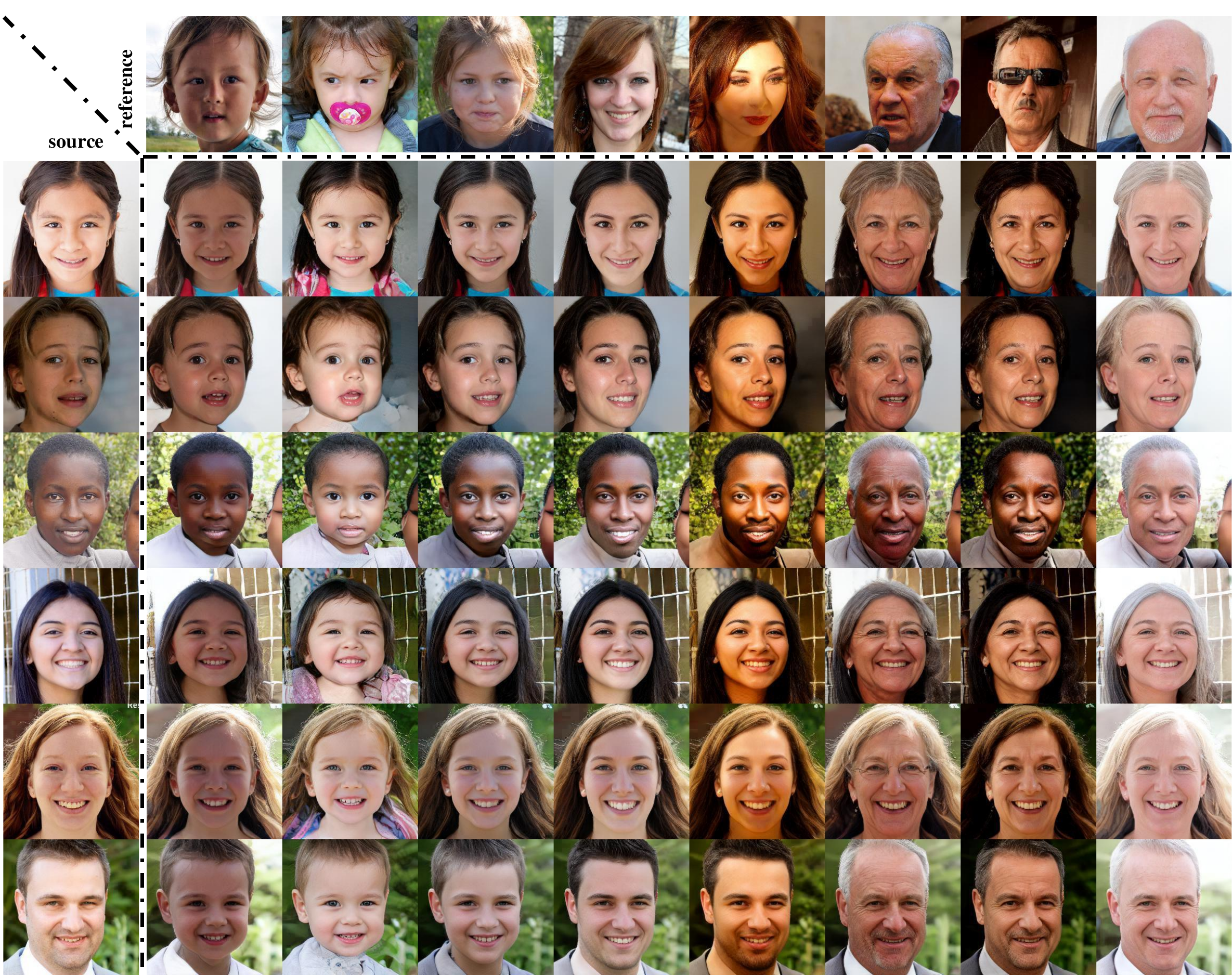}
		\caption{\textbf{Reference-based visual results on FFHQ-Aging.} The source and reference images in the first column and the first row are real images, while the rest are generated images. According to the different reference images, we make the generated images have the corresponding age domain characteristics, and have similar style to the reference image.}
		\label{fig:ref0}
	\end{figure*}
	
	\begin{figure*}
		\centering
		\includegraphics[width=0.8\textwidth]{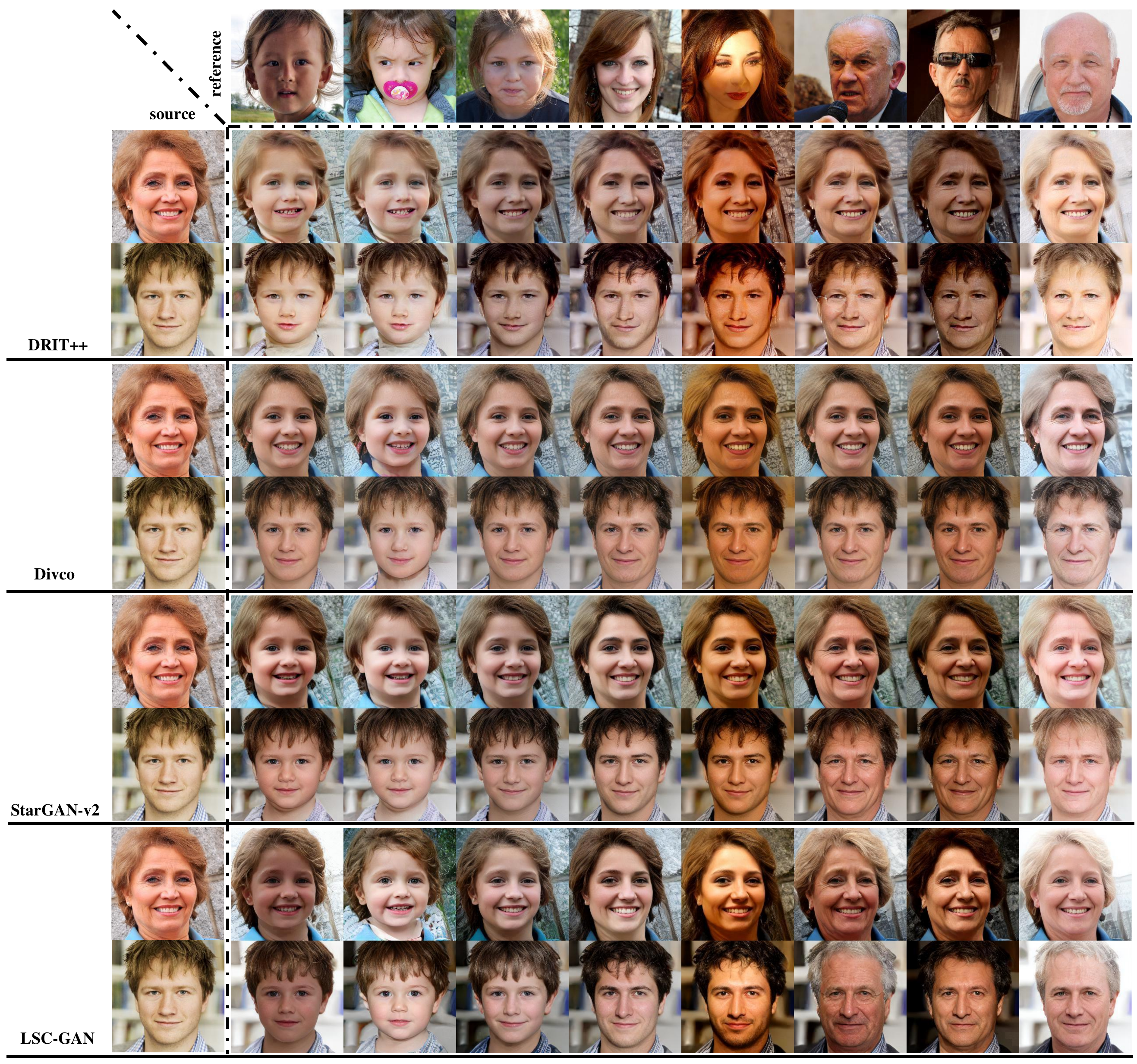}
		\caption{\textbf{Reference-based comparison on FFHQ-Aging.} Here is a visual comparison with other work.}
		\label{fig:ref1}
	\end{figure*}
	
	\begin{figure*}
		\centering
		\includegraphics[width=0.8\textwidth]{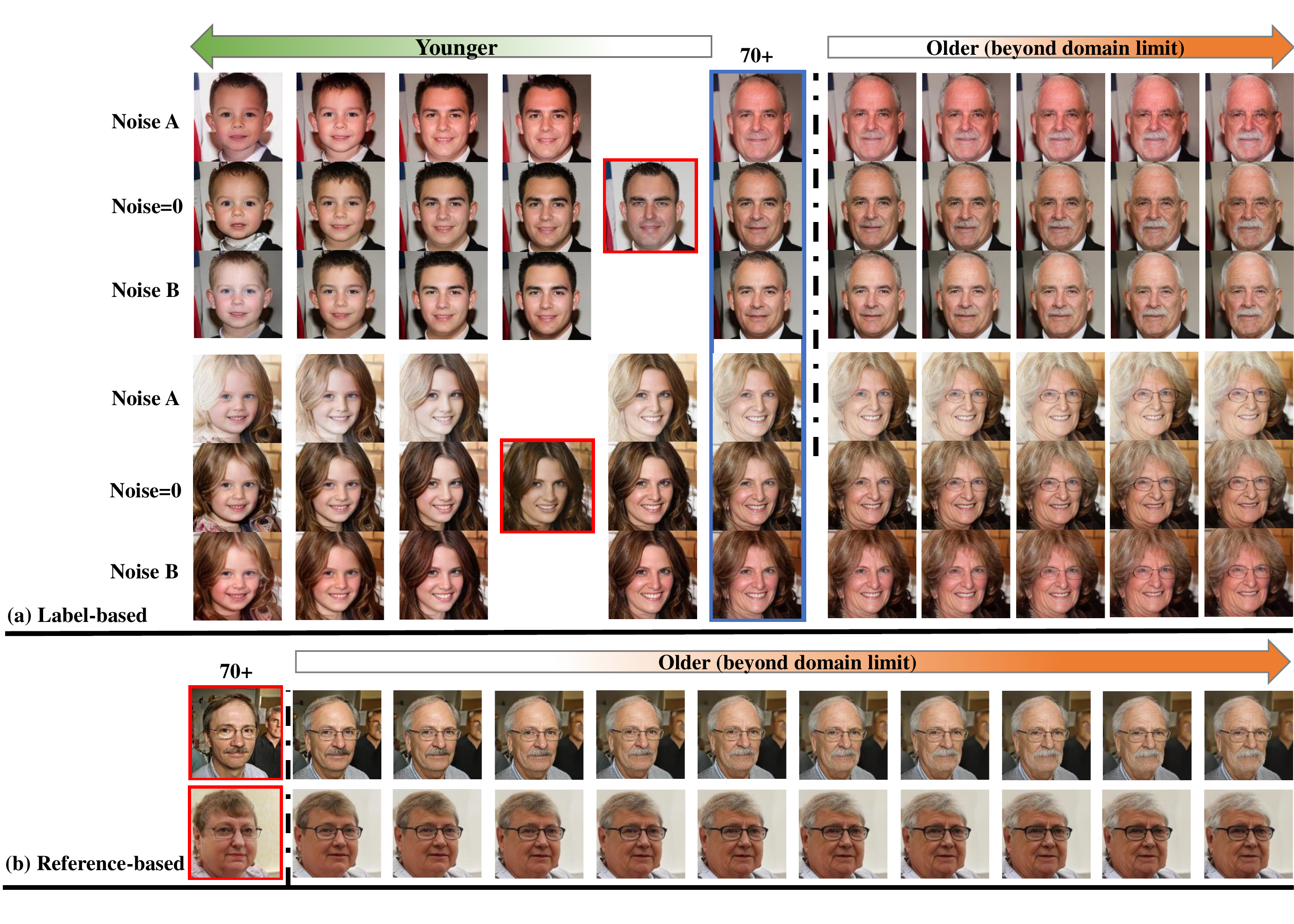}
		\caption{\textbf{Beyond domain limit.} (a) Label-based results with different noises. For the right part, we show the results of calling the module Y by 1-5 times in a row. For the left, we continuously edit the latent code by the module O, though the result in the blue box is already in the oldest domain. (b) The two faces already belong to the oldest domain. We edit them by the module O for 1-10 times, which can obviously show the older effect even beyond the maximum limit.}
		\label{fig:latent3}\vspace{-0.5cm}
	\end{figure*}
	
	\begin{figure*}
		\centering
		\includegraphics[width=0.8\textwidth]{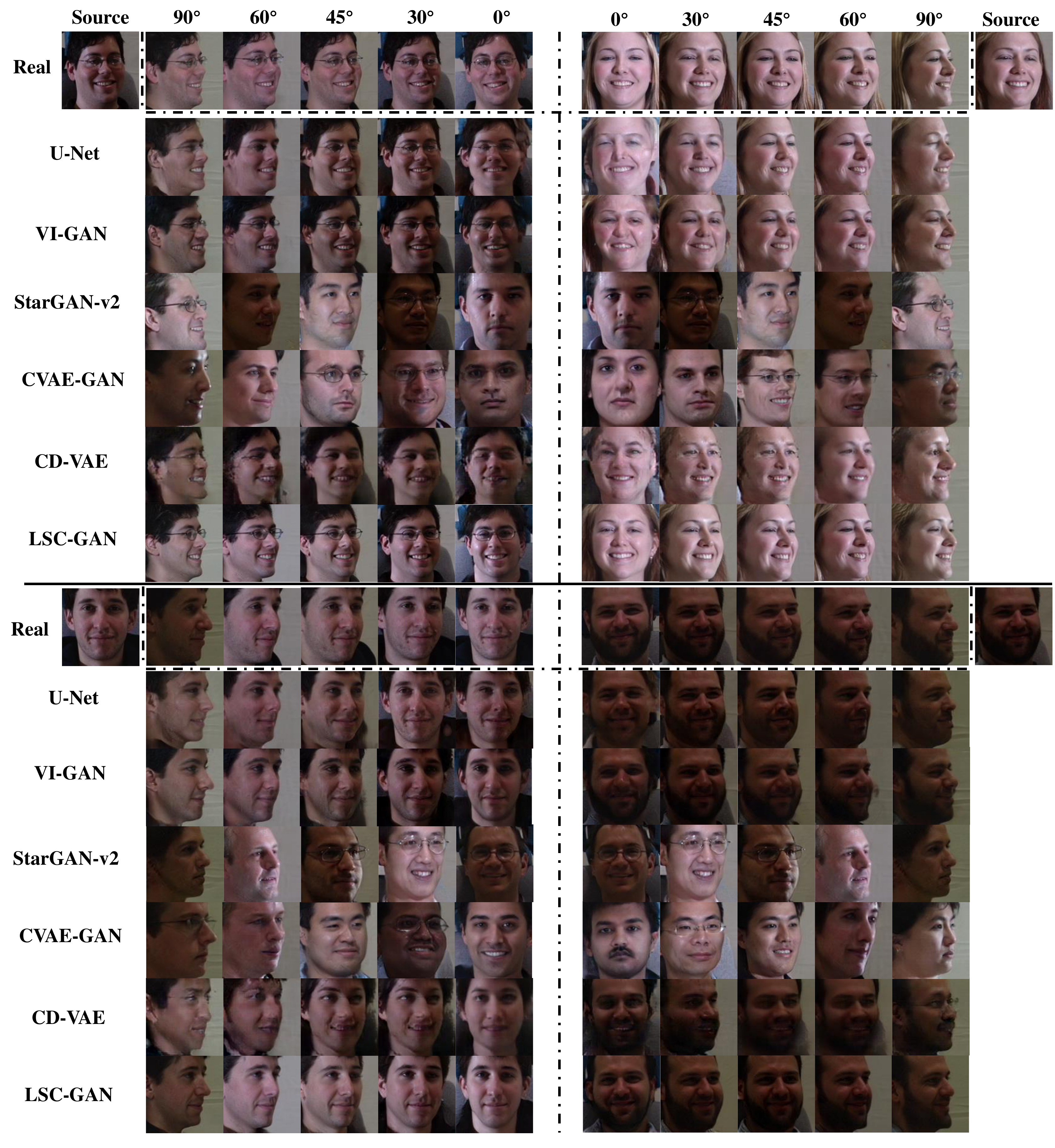}
		\caption{\textbf{Label-based comparisons on Multi-PIE.} We randomly selected 4 samples to compare the results of label-based view translation. For each sample, the first row is the ground truth, and the rest are translated images from 6 different models.}
		\label{fig:latent2}
	\end{figure*}
	
	\begin{figure*}
		\centering
		\includegraphics[width=0.8\textwidth]{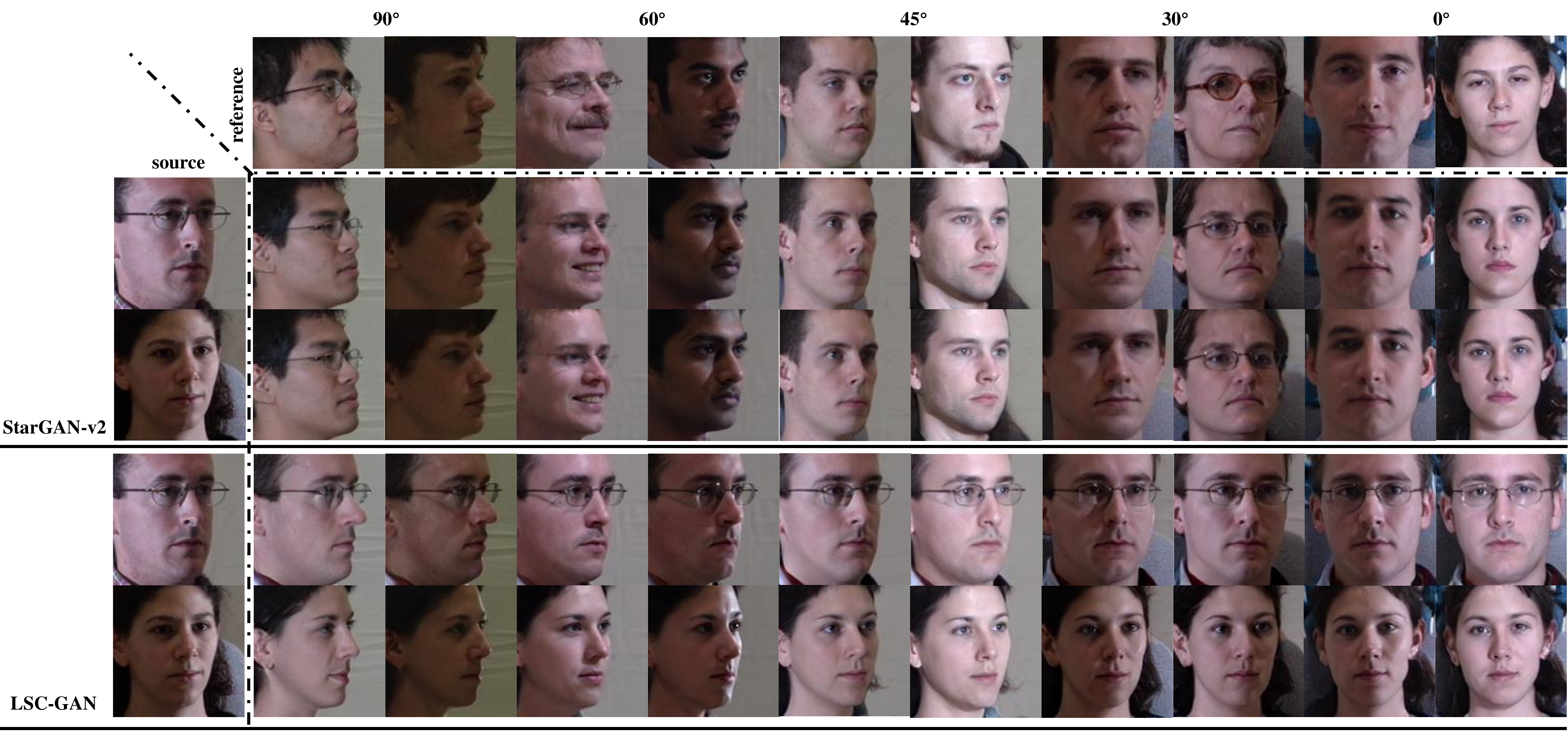}
		\caption{\textbf{Reference-based comparison on Multi-PIE.} In the first row, we randomly selected two references at each viewing angle with different lighting and identity. We compare the experimental results with StarGAN-v2.}
		\label{fig:ref2}\vspace{-0.5cm}
	\end{figure*}
	
	\section{Experiments}
	We choose two tasks to validate the proposed model for continuous domain I2I translation, which are face aging and view translation tasks. There are continuous domains in both tasks. In the former, the domains are defined by the age, while in the latter, they are determined by the geometry angle.
	\subsection{Datasets} \textbf{FFHQ-Aging.} According to the latest work LIFE \cite{or2020lifespan}, FFHQ-Aging is the best dataset so far, in terms of the amount, variety and resolution of images. So, we adopt FFHQ-Aging to evaluate our method,  which contains 70,000 images. They are categorized into 10 age groups: 0–2, 3–6, 7–9, 10–14, 15–19, 20–29, 30–39, 40–49, 50–69 and 70+. Among them, 63,000 images are used as the training set, 500 as the validation set, and 6,500 as the testing set. We resize each image to 256$\times$256. However, the confidence on the age label is not high, which may lead to inaccurate class conditional distribution modeling by the I2I model. In addition, to alleviate the problem of unbalanced sample in each group, and increase the amount of training images in each domain, we merge the groups into 6 continuous domains, 1\{0–2, 3–6\}, 2\{7–9, 10–14\}, 3\{15–19\}, 4\{20–29, 30–39\}, 5\{40–49, 50–69\}, 6\{70+\}.
	
	\textbf{Multi-PIE.} 
	We also evaluate the proposed model on the Multi-PIE \cite{gross2010multi} for the task of viewing angle translation, which contains about 130,000 images, with 13 different viewing angles, spanning from $-90^{\circ}$ to $90^{\circ}$. Each image is resized into 128$\times$128 during training. To test our hypothesis effectively, five angles (0, 30, 45, 60, 90) are used for training and testing. For details, 40,000 images are used as the training set, 2,000 as the validation set, and 8,000 as the testing set.
	\subsection{Implementation details.} Loss weights in (\ref{5}) are set as $\lambda_{cdc}=1$, $\lambda_{tam}=1$ and $\lambda_{ccc}=1$. All networks are optimized by Adam \cite{2014Adam} solver ($\beta$1= 0.5, $\beta$2= 0.999). The batch size is set to eight and the model is trained for 70K iterations. The training time is about 58 hours on a single NVIDIA GeForce RTX 3090. On networks G and D, we adopt the same network structure as StarGAN-v2. We also use the same basic residual network with IN \cite{ulyanov2016instance} as StarGAN-v2 on M, F and E. For the proposed network Y\&O, the MLP structure with residuals is adopted. The code is implemented with pytorch \cite{paszke2019pytorch}.
	\subsection{Evaluation metrics} We use Frechét inception distance (FID) \cite{NIPS2017_8a1d6947} to evaluate the visual quality, and evaluate the diversity of 
	translated images by the learned perceptual image patch similarity (LPIPS) \cite{Zhang_2018_CVPR,2012ImageNet}. For Multi-PIE, the face identity recognition network pretrained on VGGface \cite{parkhi2015deep} dataset is employed to calculate the identity accuracy of the translated image. Here the metrics for all domains are computed on the test set, and we report their averages.
	
	\begin{table}[h]
		\begin{center}
			\scalebox{1.0}{
				\setlength{\tabcolsep}{1.5mm}{
					\begin{tabular}{ l l l c| c}
						\hline
						&method && FID$\downarrow$&LPIPS(\%)$\uparrow$\\
						\hline
						&StarGAN-v2 \cite{choi2020stargan}&& 47.92$|$48.31&9.6$|$9.4\\
						&Divco \cite{liu2021divco}&& 49.00$|$56.63&14.5$|$7.3\ \ \\
						&DRIT++ \cite{lee2020drit++}&& 54.52$|$53.18&11.6$|$9.3\ \ \\
						&LIFE \cite{or2020lifespan}&& 207.86$|$-\qquad\ \ &2.7$|$-\ \ \ \\
						\hline
						&w/oFM&&44.60$|$45.05&\ 13.0$|$10.4\\
						&w/oContinuousLoss&&45.60$|$45.82&\ 12.8$|$11.5\\
						&w/oTrip&&43.85$|$44.47&\ 15.1$|$\textbf{18.9}\\
						&LSC-GAN(ours)&&\textbf{42.54}$|$\textbf{42.82}&\ \textbf{17.5}$|$17.2\\
						\hline
				\end{tabular}}
			}
		\end{center}
		\caption{
			Quantitative comparisons with state-of-the-arts on the FFHQ-Aging by FID and LPIPS. We measure on two types of synthesis. On the left of the separator "$|$" is the value of label-based synthesis, while the right is reference-based. "-" means the model can not realize the reference-based edition.}
		\label{tab:2}
	\end{table}
	
	\begin{table}[h]
		\begin{center}
			\scalebox{1.0}{
				\setlength{\tabcolsep}{1.5mm}{
					\begin{tabular}{ l l l c| c}
						\hline
						&method && quality(\%)$\uparrow$&degree(\%)$\uparrow$\\
						\hline
						&StarGAN-v2 \cite{choi2020stargan}&& 13.4&6.5\\
						&Divco \cite{liu2021divco}&& 5.1&4.8\\
						&LIFE \cite{or2020lifespan}&& 10.6&3.2\\
						&LSC-GAN(ours)&&\textbf{70.8}&\textbf{85.5}\\
						\hline
				\end{tabular}}
			}
		\end{center}
		\caption{User Study on FFHQ-Aging. We calculate and report the average of each indicator separately.}
		\label{tab:us}\vspace{-0.3cm}
	\end{table}
	
	\begin{table}[h]
		\begin{center}
			\scalebox{1.0}{
				\setlength{\tabcolsep}{1.5mm}{
					\begin{tabular}{ l l l c| c}
						\hline
						&method && FID$\downarrow$&ID(\%)$\uparrow$ \\
						\hline
						&U-Net \cite{ronneberger2015u}&& 26.50& 56.7 \\
						&CVAE-GAN \cite{bao2017cvae}&& 26.32&26.3\\
						&VI-GAN \cite{xu2019view}&& 25.04&57.1\\
						&CD-VAE \cite{yin2020novel}&& 23.95&60.4\\
						&StarGAN-v2 \cite{choi2020stargan}&& 27.20$|$23.81&0.7$|$0.4\\
						\hline
						&LSC-GAN(ours)&&\textbf{19.18}$|$\textbf{18.29}&\textbf{68.0}$|$\textbf{71.8}\\
						\hline
				\end{tabular}}
			}
		\end{center}
		\caption{Quantitative comparisons with approaches on the Multi-PIE by FID and ID accuracy. For StarGAN-v2 and DLC-GAN(ours), we measure 
			on two types of synthesis. }
		\label{tab:3}\vspace{-0.5cm}
	\end{table}
	
	\section{Results and Analysis}
	\subsection{Qualitative and Quantitative Results on FFHQ-Aging}
	According to LIFE \cite{or2020lifespan}, the traditional age translation methods, like IPCGAN, etc., cannot generate diverse results, and are often limited to the facial image captured in a specific condition, like the frontal view and the constant lighting condition. To be fair, we mainly choose StarGAN-v2 \cite{choi2020stargan} and DivCo \cite{liu2021divco}, the state-of-the-art approaches in terms of quality and diversity, for comparison. To pay tribute to the proposal of large-scale age editing, we also compare with LIFE.
	\subsubsection{Label-based synthesis}The results in Fig.\ref{fig:latent1} demonstrate that our method can produce diverse and high-quality images. The diversity is reflected in 
	intra-domain (lighting style, hair color, \emph{etc}.) and inter-domain (head size, wrinkles, \emph{etc}.). Note that although the former we show is obviously correlated with age, it does not affect the classification of domain categories. Compare with our method, the image quality of StarGAN-v2 is acceptable, but it obviously lacks the inter-domain diversity, and the face does not change except the color. DivCo has the inter-domain diversity, but still lacks intra-domain diversity, and the image quality is poor. Similarly, the generated images of DRIT++ also has artifacts, especially in the bangs. LIFE cannot generate diverse results, and it shows obvious image artifacts. In general, these methods can only synthesize the average age of each domain, while our model is able to change the age within the domain. For example, in the first domain (0-6), we can freely express all the features of 0-6 years old, but other methods can only generate features of 3 years old. 
	
	Tab.\ref{tab:2} lists FID and LPIPS of all competing methods. For LPIPS, we randomly sample 10 different noises for each test image. Then 
	translate them to the same target domain and measure the distance between the any two results. Finally, we traverse 6 target domains and get the average. For FID, we calculate it between the generated image used in calculating LPIPS and the testing set image, and calculate the average. Obviously, our method achieves the best on both indicators. In fact, it is challenging to obtain high diversity, especially the inter-domain one, while maintaining high quality. 
	
	\subsubsection{Reference-based synthesis} In Fig.\ref{fig:ref0}, we list results from 48 different pairs, consisting of 6 sources and 8 references. It can be seen that our method can naturally and accurately extract the characteristics such as the age, illumination and other styles from the reference image. However, the identity and pose of the source image are kept as much as possible. In addition, we take two sources to compare with other works, shown in Fig.\ref{fig:ref1}. StarGAN-v2 can extract the age-irrelevant factors, but still cannot synthesize different age according to the reference. Note that the first and second references belong to the first domain (0-6), but have different specific ages. The results of StarGAN-v2 only show lighting changes within the same domain. DivCo's edition has the variation on the age, but the results are obviously out of control, making the translation far from the age of the reference. In contrast, DRIT++ not only has the same problems as StarGAN-v2, but also has serious artifacts. Taking the hair color as an example, StarGAN-v2, DivCo and DRIT++ cannot change it according to the reference.

	To quantify our advantages, the FID and LPIPS are also shown in Tab.\ref{tab:2}. For LPIPS, we randomly sample reference images, and keep all of them the same for fair comparison. Our results are still the best, which proves that the proposed model can take the desired style from the reference image. 
	\subsubsection{Beyond domain limit} In the training phase, the module Y\&O is used to edit and constrain the latent code. During inference, they can still be utilized to translate the code even beyond the domain limit. 
	We show the editing effect in Fig.\ref{fig:latent3}. 
	The upper bound of the label in Fig.\ref{fig:latent3}(a) is 70+, when we continuously call the module Y, the face gradually becomes younger. Similarly, the module O changes the image into the older domain. Note that we do not generate image by the Y\&O module in the training phase, which means that the encoder M and E have not seen the code given by the Y\&O. All existing works require the same domain range during the training and inference phase. 
	But ours can translate image beyond the training range, and continue to modify the facial age. These images still have the high quality. The Y\&O module can also perform the reference-based translation, shown in Fig.\ref{fig:latent3}(b). Although the first column is labeled as the oldest, the module O still edits the latent code for the older age.
	\subsubsection{Ablation study} In Tab.\ref{tab:2}, we show the effect of each proposed item in detail by removing it from the original model. 
	\textbf{w/oFM} represents the same structure as StarGAN-v2 for label-based synthesis without the module F and M. All the loss functions and the Y\&O module are still retained. Both FID and LPIPS are getting worse, which means that the structure of StarGAN-v2 is not suitable for the continuous domain translation. 
	It is irrational to directly map noises or reference images into the style codes reflecting the domain interval. Since the style code tends to be close to the center of the domain, it reduces the diversity of translated image. In contrast, our structure first maps a constant into the latent domain code that characterizes the center of each domain interval, and then uses noises or reference images to give the diverse styles, which not only expresses the age within the domain, but also prevents the mode collapse. 
	\textbf{w/oContinuousLoss} removes the $L_{cdc}$ and $L_{ccc}$ in Equation.(\ref{1}) and (\ref{4}). Since they have similar functions, so we bind them together for ablations. Poor quality and low LPIPS show that they are indispensable in establishing the link among domains. 
	\textbf{w/oTri}
	removes the $L_{tam}$ in Equation.(\ref{4}). The performance in Tab.\ref{tab:2} also drops significantly. Actually, the $L_{tam}$ constrains distances of the latent codes among different domains, which helps the module Y\&O to fit the age trend. 
	\subsubsection{User Study} We conduct user study in the form of a questionnaire.The subjects are provided with 20 sets of generated images, and are required to select the most suitable one from each set. We ask the subjects which result has the best quality and which has the maximum degree of translation. In detail, a group of translated results for the same source image are listed for the users, and these images come from 4 different models of StarGAN-v2, DivCo, LIFE and LSC-GAN. The statistical results are listed in Tab.\ref{tab:us}. Our model achieves the best on both quality and translation degree, which shows that from a human perspective, the images edited by our network are more in line with human aesthetics.
	\subsection{Qualitative and Quantitative Results on Multi-PIE} We also carry out the view translation on Multi-PIE. Different from face aging, the ground truth of the target view is known, but we do not use it and keep the task unsupervised during training. We choose the model U-Net \cite{ronneberger2015u}, CVAE-GAN \cite{bao2017cvae}, and the advanced works VI-GAN \cite{xu2019view}, CD-VAE \cite{yin2020novel} for comparison. Note that it is unfair for our LSC-GAN to measure the SSIM or MSE between the ground truths and results, because our model can synthesize diverse results which are not necessarily the same as ground truths. \subsubsection{Label-based synthesis}
	Fig.\ref{fig:latent2} shows the visual comparisons. StarGAN-v2 and CVAE-GAN obviously cannot maintain the identity of the source image. The image quality of U-Net and CD-VAE is poor, with serious artifacts. VI-GAN performs well in some viewing angles, but overall it is still inferior to us, especially in image quality. Our LSC-GAN gives the satisfactory results, and it accurately translates the source images into different viewing angles. In Tab.\ref{tab:3}, we measure the FID and ID accuracy to further prove the advantages of our model. There is no doubt that we achieve the best on both metrics. Note that we do not adopt any loss that maintains identity, but our model can still keep it well under different views. This proves that the proposed method can model the latent style code to establish and fit the continuous trends. 
	\subsubsection{Reference-based synthesis} Less work can successfully complete the reference-based viewing angle translation. As shown in Fig.\ref{fig:ref2}, our model can not only change the view angle according to the reference and maintain the source identity, but also can realistically synthesize the lighting environment in the reference image. 
	Compared with StarGAN-v2, although it can also capture the illumination condition in the reference, the identity is seriously affected by the reference, and the source image is ignored. We also measure the quantitative results in Tab.\ref{tab:3}, which shows the proposed LSC-GAN can give the best result. This once again proves the superiority of our model for modeling the latent style codes and their relation for continuous domain translation. 
	
	\section{Conclusion}
	This paper investigates the continuous I2I translation task. We propose the LSC-GAN for both label- and reference-based translation, which intentionally builds the latent style codes for each domain interval, and models their relation by the designed code editing modules. These modules are able to translate the style code from one domain to its neighbours, so they can recursively change the style code from source to target. Moreover, the LSC-GAN achieves the diverse synthesis by explicitly incorporating the sampling noises and the references into the latent style code. Extensive experiments on two datasets for continuous domain I2I demonstrate the effectiveness of the proposed model.
	
	\ifCLASSOPTIONcaptionsoff
	\newpage
	\fi
	
	{\tiny
		\bibliographystyle{ieee_fullname}

	}
\end{document}